# Extending Universal Intelligence Models with Formal Notion of Representation

Alexey Potapov, Sergey Rodionov

AIDEUS, Russia
{potapov,rodionov}@aideus.com

**Abstract.** Solomonoff induction is known to be universal, but incomputable. Its approximations, namely, the Minimum Description (or Message) Length (MDL) principles, are adopted in practice in the efficient, but non-universal form. Recent attempts to bridge this gap leaded to development of the Representational MDL principle that originates from formal decomposition of the task of induction. In this paper, possible extension of the RMDL principle in the context of universal intelligence agents is considered, for which introduction of representations is shown to be an unavoidable meta-heuristic and a step toward efficient general intelligence. Hierarchical representations and model optimization with the use of information-theoretic interpretation of the adaptive resonance are also discussed.

**Key words:** Universal Agents, Kolmogorov Complexity, Minimum Description Length Principle, Representations

## 1 Introduction

The idea of universal induction and prediction on the basis of algorithmic information theory was invented a long time ago [1]. In theory, it eliminates the fundamental problem of prior probabilities, incorrect solutions of which result in such negative practical effects as overlearning, overfitting, oversegmentation, and so on. It would be rather natural to try to develop some models of universal intelligence on this basis. However, the corresponding detailed models were published only relatively recently (e.g. [2]). Moreover, the theory of universal induction was not popular even in machine learning. The reason is quite obvious – it offers incomputable methods, which additionally require training sets of large sizes in order to make good predictions.

Unsurprisingly, such more practical alternatives as the Minimum Description Length (MDL) or the Minimum Message Length (MML) principles became much more popular. These principles help developers to considerably improve performance of machine learning and perception methods, but still they neither completely solve the problem of prior probabilities nor allow for universal machine learning systems.

Of course, the universal intelligence models inherit the same drawbacks as the universal prediction. Namely, computational intractability is even more considerable here. Optimality of the models is proven up to some constant slowdown factor that can be very large. This slowdown can be eliminated via self-optimization [3], but its time for unbiased intelligence can also be very large. Consequently, most researchers consider universal models as possibly interesting, but pure abstract tools.

At the same time, practical success of the MDL principle and its counterparts implies that there is a way toward a realistic implementation of universal induction. However, there is still a very large gap to be bridged. Indeed, applications of the MDL principle rely on hand-crafted heuristic coding schemes invented by developers for each specific task. These schemes specify algorithmically incomplete model spaces with large inductive bias resulting only in weakly learnable systems.

In order to bridge this gap, the notion of representation was recently formalized within the algorithmic information theory, and the Representational MDL (RMDL) principle was introduced [4]. This principle can be used to estimate quality of decomposition of the task of model construction for some large data series into relatively independent subtasks. Residual mutual information between these subtasks can be taken into account by adaptive resonance models, which also have the information-theoretic formalization [5].

In this paper, we consider application of the RMDL principle as an unavoidable meta-heuristic for the model of the universal algorithmic intelligence. Only one heuristic is not enough to achieve efficient universal intelligence, but it makes this goal a little bit closer.

## 2  Background

The model of intelligence as some sort of search for the best chain of actions was the first one adopted in the AI field. It can be applied for solving any problem, but only in the case of known determined settings and unlimited computational resources. Universal Solomonoff induction/prediction affords an opportunity to extend this model on the cases of arbitrary (computable) unknown environments. However, the problem of computational resources remains and becomes more complicated. Moreover, unbiased universal agent may need a lot of time to acquire necessary information about the world to become able to secure own survival even possessing infinite computational resources. Because speeding up the search for chains of actions can also be treated as learning, the induction problem should be considered in the first place.

Solomonoff induction relies on the notion of algorithmic probability, which is calculated for a binary string α as:

$$P_U(\alpha) = \sum_{p:U(p)=\alpha} 2^{-l(p)}, \qquad (1)$$

where $U$ is some Universal Turing Machine (UTM), and $p$ is its program with length $l(p)$ that produces the string α being executed on the UTM $U$.

Probabilities $P_U(\alpha)$ are referred to as the universal prior distribution. Why are they universal? The basic answer to this question rests on the fact that any universal machine $U$ can be emulated on another universal machine $V$ by some program $u$: for any $p$, $V(up)=U(p)$. Consequently,

$$P_U(\alpha) = \sum_{p:U(p)=\alpha} 2^{-l(p)} = 2^{l(u)} \sum_{p:V(up)=\alpha} 2^{-l(up)} \leq 2^{l(u)} P_V(\alpha), \qquad (2)$$

and similarly $P_V(\alpha) \leq 2^{l(v)} P_U(\alpha)$.

This implies that difference between the algorithmic probabilities of arbitrary string α on any two UTMs is not more than some multiplicative constant independent of α. Given enough data, likelihood will dominate over the difference in prior probabilities, so the choice of the UTM seems to be not too crucial.

However, the amount of necessary additional data can be extremely large in practice. One can still refer to the algorithmic probabilities as universal priors, because no other distribution can be better in arbitrary unknown environment. Universality of this distribution simply means that it is defined on the algorithmically complete model space (any algorithm has non-zero probability and can be learned), and models are naturally ordered by their complexity (it is impossible to specify such universal machine that reverts this order).

Apparently, the universal agent based on the algorithmic probability (such as AIξ [2]) may require executing many actions to make history string long enough to neutralize influence of the arbitrarily selected *U*. And no unbiased intelligence can perform better.

However, we don't want our universal agent to be absolutely unbiased. Quite the contrary, we do want it to be universal, but biased towards our world. In this context, dependence of the algorithmic probabilities on the choice of UTM appears to be very useful in order to put any prior information and to reduce necessary amount of training data. This idea was pointed out by different authors [6, 7]. It is also said [8] that the choice of UTM can affect the "relative intelligence of agents".

Unfortunately, no universal machine can eliminate necessity for exhaustive search for algorithms that produce the whole agent's history. At the same time, the pragmatic MDL principle is applied to algorithmically incomplete model spaces specified by hand-crafted coding schemes, which allow for efficient non-exhaustive search procedures. Of course, it is unacceptable to replace UTMs with Turing-incomplete machines as the basis of the universal intelligence. Can this intelligence apply the MDL principle in the same way as we do?

## 3   Representational MDL Principle

The minimum description length principle states that the best model of the given data source is the one which minimizes the sum of
   – the length, in bits, of the model description;
   – the length, in bits, of data encoded with the use of the model.

In theory, this principle is based on the Kolmogorov (algorithmic) complexity $K_U(\alpha)$ that is defined for some string α as:

$$K_U(\alpha) = \min_p [l(p) | U(p) = \alpha]. \qquad (3)$$

The MDL principle is derived from the Kolmogorov complexity if one divides the program $p$ for UTM $p=\mu\delta$ into the algorithm itself (the regular component of the model) $\mu$ and its input data (the random component) $\delta$:

$$K_U(\alpha) = \min_p [l(p) \,|\, U(p) = \alpha] = \min_{\mu\delta}[l(\mu\delta) \,|\, U(\mu\delta) = \alpha] = \min_\mu \min_\delta [l(\mu) + l(\delta) \,|\, U(\mu\delta) = \alpha] = \min_\mu \left[ l(\mu) + \min_\delta [l(\delta) \,|\, U(\mu\delta) = \alpha] \right] = \min_\mu [l(\mu) + K_U(\alpha \,|\, \mu)]. \quad (4)$$

Here, $K_U(\alpha \,|\, \mu) = \min_\delta [l(\delta) \,|\, U(\mu\delta) = \alpha]$ is the conditional Kolmogorov complexity of $\alpha$ given $\mu$. Consequently, the equation

$$\mu^* = \arg\min_\mu [l(\mu) + K(\alpha \,|\, \mu)] \quad (5)$$

gives the best model via minimization of the model complexity $l(\mu)$ and the model "precision" $K(\alpha \,|\, \mu) = l(\delta)$, where $\delta$ describes deviations of the data $\alpha$ from the model $\mu$. This equation becomes similar to the Bayesian rule, if one assumes $-\log_2 P(\mu) = l(\mu)$ and $-\log_2 P(\alpha \,|\, \mu) = K(\alpha \,|\, \mu)$.

The MDL principle differs from the algorithmic probability in two aspects. The first one consists in selection of a single model. It can be useful in communications between intelligent agents or for reducing the amount of computations [9], but in general the MDL principle is a rough approximation of the algorithmic probability.

The second aspect consists in adopting the two-part coding. In practice, it helps to separate regular models from noise. This separation can be considered as a useful heuristic, but it is somewhat arbitrary within the task of model selection. In any case, Kolmogorov complexity is also incomputable. Thus, we still need to bridge the gap between the theoretical MDL principle and its practical applications. This is done (to some extent) within the Representational MDL principle.

The main idea here is that machine perception and machine learning methods are applied in practice to mass problems (sets of separate, individual problems of some classes). For example, any image analysis method is applied to different images independently searching for separate image descriptions in a restricted model space. On the contrary, the universal intelligence agent enumerates algorithms producing the whole history string. Let this history consists of a number of substrings (e.g. images) $\alpha_1 \alpha_2 \ldots \alpha_n$. If the agent tries to compute individual Kolmogorov complexities (or algorithmic probabilities) of these strings, the result in the most cases will be poor:

$$\sum_{i=1}^n K_U(\alpha_i) \gg K_U(\alpha_1 \alpha_2 \ldots \alpha_n), \quad (6)$$

because these substrings normally contain a lot of mutual information. This mutual information (let it be denoted by $S$) should be removed from descriptions of individual data strings, and should be considered as prior information in corresponding subtasks of analysis of individual substrings. This implies usage of the conditional Kolmogorov complexities $K(\alpha_i \,|\, S)$. Indeed, one can expect that

$$K_U(\alpha_1\alpha_2...\alpha_n) \approx \min_S \left( l(S) + \sum_{i=1}^{n} K_U(\alpha_i \mid S) \right) << \sum_{i=1}^{n} K_U(\alpha_i). \qquad (7)$$

Since $S$ can be interpreted as an algorithm (some program for UTM), which produces any given data string from its description, the algorithm $S$ precisely fits the verbal notion of representation formulated by David Marr [10]. The notion of representation is treated in the same way in the papers on AGI (e.g. "internal representation interprets input reconstructing it" [11]). Therefore, the following more strict definition can be given [4].

**Definition.** The program $S$ for UTM $U$ is called *representation* of the collection of data strings (e.g. images) $A = \{\alpha_1,...,\alpha_n\}$, if $(\forall \alpha \in A)(\exists \mu, \delta \in \{0,1\}^*) U(S\mu\delta) = \alpha$. The string $\mu\delta$ is called *description* of $\alpha$ within the representation $S$. This description consists of the regular $\mu$ and the random $\delta$ components.

Consequently, the RMDL principle states that 1) the best model of the data string *within given representation* is the model, for which the sum of the length of the model and the length of this data string described with the use of this model is minimal; 2) the best representation of the collection of the data strings is the representation, for which the sum of the length of the representation and the summed length of the minimal descriptions of these data strings within the representation is minimal.

When we consider any practical image analysis method, it uses some representation of images. This representation specifies an inductive bias similar to that specified by the choice of the UTM in algorithmic complexity or probability. However, the universal agent is based on the single UTM, while representations can differ for different sensor modalities or even for different elements of the same modality, they can be Turing-incomplete, and they can be learned and changed during lifetime.

It is interesting to note that for any two UTMs $U$ and $V$ and for any representation $S$ for $U$ there exists the equivalent representation $S'$ for $V$ such that $K_U(\alpha \mid S) = K_V(\alpha \mid S')$ for any $\alpha$. Indeed, it is obvious for $S'=uS$, where $u$ emulates $U$ on $V$. Thus, the choice of UTM influences on the representation construction, but not on the model selection within equivalent representations. Thus, we will write $K_S(\alpha)$ instead of $K_U(\alpha \mid S)$, and $K_S(\alpha \mid \mu)$ instead of $K_U(\alpha \mid S\mu)$.

It should be pointed out that the RMDL principle is not just an extension of the two-part coding to a "three-part" coding. Any three- (or more) part coding of an individual string could be re-structured to the two-part coding scheme [9], but $S$ and $\mu$ in the RMDL principle cannot be united, because $S$ describes the problem class, while $\mu$ describes its instance.

It is also interesting to note that the idea of deep learning architectures [12] arose from the fact that complexity of some models is exponentially larger within shallow representations than within deep representations. The RMDL principle allows for much more detailed analysis of the representation efficiency.

## 4 Hierarchical Representations and Adaptive Resonance

Separate descriptions of substrings even within a good representation will still contain some mutual information (large-scale regularities in the initial string). Thus, if one has the string α divided into the substrings $α_1α_2…α_n$, and the descriptions $μ_iδ_i$ are independently constructed for each substring, it is natural to try to compress the string $μ=μ_1μ_2…μ_n$ (deltas can be ignored on the next level of description since they are interpreted as noise within the RMDL principle). This string can still be very long, so one would like to divide μ into larger substrings (or to group $μ_i$) and to describe these substrings within some higher-level representation. Resulting models (regular parts of descriptions) can be further compressed, and so on.

Specific division of the string into substrings can be unknown a priori and can be considered as a part of a model. For example, borders of word and sentence segments in speech signals are not known. Images also should be segmented into some regions, which content can be described almost independently. For now, we can ignore the structure of these models and use only whole strings.

That is, at first the model $μ^{(1)}$ is constructed for the string α within the representation $S^{(1)}$. Then, the model $μ^{(2)}$ is constructed for the string $μ^{(1)}$ within some higher-level representation $S^{(2)}$, and so on up to some level of abstraction $m$:

$$μ^{(1)} = \arg\min_{μ}[l(μ) + K_{S^{(1)}}(α\,|\,μ)],$$

$$μ^{(i+1)} = \arg\min_{μ}[l(μ) + K_{S^{(i+1)}}(μ^{(i)}\,|\,μ)]. \tag{8}$$

The total description length (an approximation of Kolmogorov complexity) of the string α can be calculated as:

$$L_{S^{(1)},…,S^{(m)}}(α) = K_{S^{(1)}}(α\,|\,μ^{(1)}) + \sum_{i=2}^{m-1} K_{S^{(i)}}(μ^{(i)}\,|\,μ^{(i+1)}) + l(μ^{(m)}), \tag{9}$$

where $K_{S^{(i)}}(μ^{(i)}\,|\,μ^{(i+1)}) = l(δ^{(i+1)})$.

It can be seen that sequential construction (8) of models of higher levels of abstraction is not the same as minimization of the total description length (9). Indeed, one should search for the models on all levels of abstraction simultaneously in order to get the optimal result (9). However, such the exhaustive search is computationally expensive. The sequential model construction is much more practical, but much less robust, because it is bottom-up and greedy.

Here, one can adopt Grossberg Adaptive Resonance Theory. Some subsets of models should be considered on each level of abstraction, and models on different levels should support or suppress each other. Such models remain, for which resonance is established. Qualitative expression of support values can be derived from the RMDL principle in the form of equation (9), so it can be used in the information-theoretic formalization of the Adaptive Resonance Theory [5].

Hierarchical decomposition of a problem into slightly dependent sub-problems, construction of their separate solutions, and adaptive correction of these solutions in accordance with the whole problem can be considered as almost universal meta-heuristic.

## 5  Adoption of the RMDL Principle in Universal Algorithmic Intelligence

The opinion that representations should be incorporated into the models of general intelligence has been already stated [13, 14]. However, representations are usually implemented only in the form of prior information expressed in a special design of programming language. Besides insufficiency of strict quantitative analysis of representation quality, the main restriction here is absence of decomposition of the model construction task.

On the other hand, necessity of decomposition is also realized. In particular, importance of chunks and possibility to solve tasks only of small Kolmogorov complexity are noted [7, 15, 16]. The RMDL principle can strictly account for both these aspects.

Consider the universal intelligent agent based on the algorithmic probability. We will use Hutter's AIξ model for convenience in order to skip unnecessary detailed descriptions of less known models. The AIξ agent is intended to maximize the total reward choosing its actions [2]:

$$y_k = \arg\max_{Y_k} \max_{p:U(px_{<k})=y_{<k}Y_k} \sum_{q:U(qy_{<k})=x_{<k}} 2^{-l(q)} V_{km_k}^{pq}, \quad (10)$$

where $y_{<k}$ is the string of agent's actions till the time moment $k$, and $x_{<k}$ is the string of sensory history (including reward signals); $p$ are possible agent's policies consistent with the history, and $q$ are possible algorithmic models of the environment; $V_{km_k}^{pq}$ is the expected future reward summed in the $[k, m_k]$ time interval executing algorithms $p$ and $q$ on the UTM $U$.

The formal notion of representation can be almost straightforwardly applied to the agent's inputs $x_{<k}$. Although the RMDL principle can be extended from Kolmogorov complexity to algorithmic probability, we will use its basic version for the sake of simplicity (differences between Kolmogorov complexity and algorithmic probability are discussed in our companion paper). If one uses only one best model $q_{opt}$, the equation (10) can be rewritten:

$$y_k = \arg\max_{Y_k} \max_{p:U(px_{<k})=y_{<k}Y_k} V_{km_k}^{pq_{opt}}, \quad q_{opt} = \arg\min_{q:U(qy_{<k})=x_{<k}} l(q), \quad (11)$$

To apply the RMDL principle, one should decompose $q_{opt}$ into some set of (nearly) independent models $q_i$ conditioned by some representation $S$ for the segmented history $x_{<k} = x_{m_1+1:m_2}...x_{m_{n-1}+1:m_n}$, where $m_1=0$ and $m_n=k$: $U(Sq_i y_{<k}) = x_{m_i+1:m_{i+1}}$ (however, it should be noted that this form of decomposition/segmentation is not universal).

In this case, $q_i$ can be sought independently. If $l(q) \approx l(S) + l(q_1) + \ldots + l(q_n)$, the complexity of the full task will be $2^{l(S)} \prod 2^{l(q_i)}$, while the complexity of the decomposed task will be $2^{l(S)} \sum 2^{l(q_i)}$ that is much smaller. One can also divide $q_i$ into the model $\mu_i$ and noise $\delta_i$ further simplifying the search problem. However, in order to calculate $V_{km_k}^{pq}$ it is necessary to predict future values $x_{k:m_k}$ of the input. This is impossible if induction is aborted after construction of the set of decomposed model $\{q_i\}$. If $q_i$ are really independent, they are unpredictable. However, this is not the case in reality. Thus, one should construct a higher level model, which produces the sequence $q_{1:n}$, and extrapolates it. A number of intermediate levels of the representation can be introduced, and the hierarchical model can be optimized with help of adaptive resonance as it was described in the previous section.

Another difference from the pure RMDL principle here is that the environment model $q$ takes agents actions $y_{<k}$ as input. Should the whole history of actions be taken for each partial model $q_i$? Probably, no. Here, one can think about representations for action history.

It is attractive to try to decompose the program $p$ in the same way as it was done for the program $q$. However, there is a huge difference between these programs. The program $q$ is used as the environment model in predicting the inputs $x$. However, the agent doesn't need to predict own actions, since they can be chosen directly:

$$y_k = \arg\max_{Y_k} \max_{Y_{k+1:m_k}} V_{km_k}^{q_{opt}(y_{<k} Y_{k:m_k})} . \quad (12)$$

This form of search is even less computationally expensive, because action chains $Y_{k:m_k}$ have bounded complexity, while programs $p$ can have arbitrarily large complexity. Thus, there is no sense to enumerate all programs $p$ and to decompose them. However, search in the space of all possible action chains is still too computationally expensive. It is clear that any simplification of this exhaustive search should be done very carefully in order to avoid substantial limitations of the agent's universality.

The notion of representation can still be useful here. One can imagine some generalized actions, which can be introduced as some combinations of elementary actions, or even as small programs $p_i$. These generalized actions will be useful only in the case, when the total number of chains of these actions is not larger than the total number of chains of elementary actions (this condition can be expressed also in probabilistic terms for stochastic search). Thus, variety of generalized actions will be smaller, and their introduction can be formally grounded only on the base of a criterion that takes computational costs of the search strategy into account. Such criterion is now absent, and possibility to mathematically introduce representations for actions can be proposed only as an idea.

It is interesting to note that if generalized actions are enumerated, one can consider models of the environment that accept chains of these generalized actions as input:

$$p_k = \arg\max_{P_k} \max_{P_{k+1:m_k}} V_{km_k}^{q_{opt}(p_{<k} P_{k:m_k})}, \quad q_{opt} = \arg\min_{q:U(qp_{<k})=x_{<k}} l(q) . \quad (13)$$

Indeed, humans rarely predict explicit reaction of the environment on their each very elementary action. At the same time, generalized actions $p_i$ can also accept generalized input strings $q_i$. Indeed, we say "take the apple" or "open the door". That is, representations for sensory data (including generalized rewards) and actions are interconnected. Search in the space of generalized entities can be greatly simplified (but representations should be still constructed using the Turing-complete space). This approach can be used to gradually introduce advanced representations as priors for efficient generally intelligent agents starting from low-level representations for raw data and elementary actions and finishing with knowledge representations.

## 6  Conclusions

The notion of representation is extremely useful for almost all cognitive functions. However, it is rarely defined strictly enough. The necessary formal definition was recently given jointly with the Representational MDL principle, which is derived from decomposition of Kolmogorov complexity. In this paper, we discussed possibility to extend the model of universal algorithmic intelligence (namely AIξ). We showed that this principle can be rather naturally incorporated into this model making it somewhat closer to efficient artificial general intelligence. Information-theoretic criteria of quality of representations and models can be used for consequently constructing more optimal methods of machine perception and learning, including multi-level systems with adaptive resonance.

However, the RMDL principle only partially solves the problem of quality of representations in the models of universal algorithmic intelligence. It was initially introduced for such tasks, which decomposition is defined a priori (e.g. a computer vision system should analyze images independently), and representations are needed in order to decrease negative effects of this decomposition. However, there is no given decomposition of the task of prediction in the case of the universal agent. Decomposition is necessary for reducing computational complexity, but it leads to increase of algorithmic (Kolmogorov) complexity of environment models. Thus, representations trade computational complexity for algorithmic complexity. Apparently, the RMDL principle based on Kolmogorov complexity is only a particular case of constant computational complexity. In future, generalized RMDL principle should be developed based on Levin complexity (e.g. defined in [17]). Representations for Levin complexity can help to strictly account for the bias in complexity of models, which are used many times in descriptions of different data segments or executed many time during prediction and sequential decision making. Another open problem consists in formalization of representations not only for sensory input, but also for actions. We believe that such formalization can help to develop a theory of efficient self-optimization.